\documentclass[runningheads]{llncs}
\usepackage[T1]{fontenc}
\usepackage{graphicx}
\usepackage{hyperref}
\usepackage{booktabs}
\usepackage{tabularx}
\usepackage{listings}
\usepackage{amsmath} 
\usepackage{nicefrac,xfrac}
\usepackage{tabularx}
\usepackage{cleveref}
\Crefname{figure}{Fig.}{Figs.}
\usepackage{algorithm}
\usepackage[noend]{algpseudocode}
\usepackage{makecell,booktabs}
\usepackage{amsfonts} 
\usepackage[most]{tcolorbox}
\usepackage[misc]{ifsym}

\usepackage{todonotes} %
\newcommand{\corr}{(\Letter)}
\NewDocumentCommand{\rot}{O{45} O{1em} m}{\makebox[#2][l]{\rotatebox{#1}{#3}}}
\usepackage{mwe}
\usepackage{xcolor}

\newcommand{\ours}{RED}
\newcommand{\per}{Permute}
\newcommand{\classic}{Classic}
\newcommand{\hyper}{Hyper Parameter Grid}
\newcommand{\fitting}{Fitting}
\newcommand{\cvgp}{CVGP}
\newcommand{\seeded}{Seeded GPLearn}
\MakeRobust{\Call}

\newcommand{\preparecomment}[1]{{\color{darkgray}$\triangleright$ #1}}

\algnewcommand{\LineComment}[1]{\State\preparecomment{\textit{#1}}}

\algnewcommand\algorithmicglobal{\textbf{Global:}}
\algnewcommand\Global{\item[\algorithmicglobal]}

\algnewcommand\algorithmicforeach{\textbf{for each}}
\algdef{S}[FOR]{ForEach}[1]{\algorithmicforeach\ #1\ \algorithmicdo}

\algblockdefx[Class]{Class}{EndClass}%
[2]{\textbf{class} \Call{#1}{#2}}%
{\textbf{end class}}

\algnotext{EndClass} %
\newcolumntype{Y}{>{\centering\arraybackslash}X}
\newcolumntype{R}{>{\raggedleft\arraybackslash}X}
\begin{document}

\title{%
Prompting Neural-Guided Equation Discovery Based on Residuals}

\titlerunning{Prompting Neural-Guided Equation Discovery Based on Residuals}

\author{
Jannis Brugger\inst{1,2}\orcidID{0000-0002-7919-4789} \corr \and
Viktor Pfanschilling\inst{1,5}\orcidID{0000-0002-4752-9003} \and
David Richter \inst{1}\orcidID{0000-0000-0000-0000}
Mira Mezini\inst{1,2,3}\orcidID{0000-0001-6563-7537} \and
Stefan Kramer\inst{4}\orcidID{0000-0003-0136-2540}
}

\authorrunning{J. Brugger et al.}

\institute{
  Technical University of Darmstadt, 64289 Darmstadt, Germany
  \email{jannis.brugger@tu-darmstadt.de}
\and
  Hessian Center for Artificial Intelligence (hessian.AI), 64293 Darmstadt, Germany
\and National Research Center for Applied Cybersecurity ATHENE
\and
  Johannes Gutenberg-Universität Mainz, 55128 Mainz, Germany
\and
  German Research Center for Artificial Intelligence, 67663 Kaiserslautern, Germany
}
\renewcommand{\floatpagefraction}{.8}

\maketitle              %

\begin{abstract}
Neural-guided equation discovery systems use a data set as prompt and predict an equation that describes the data set without extensive search. However, if the equation does not meet the user's expectations, there are few options for getting other equation suggestions without intensive work with the system.
To fill this gap, we propose Residuals for Equation Discovery (RED), a post-processing method that improves a given equation in a targeted manner, based on its residuals.
By parsing the initial equation to a syntax tree, we can use node-based calculation rules to compute the residual for each subequation of the initial equation. It is then possible to use this residual as new target variable in the original data set and generate a new prompt.
If, with the new prompt, the equation discovery system suggests a subequation better than the old subequation on a validation set, we replace the latter by the former. 
RED is usable with any equation discovery system, is fast to calculate, and is easy to extend for new mathematical operations.
In experiments on 53 equations from the Feynman benchmark, we show that it not only helps to improve all tested neural-guided systems, but also all tested classical genetic programming systems.

\end{abstract}

\begin{figure}[t]
\begin{center}
\includegraphics[width=\linewidth]{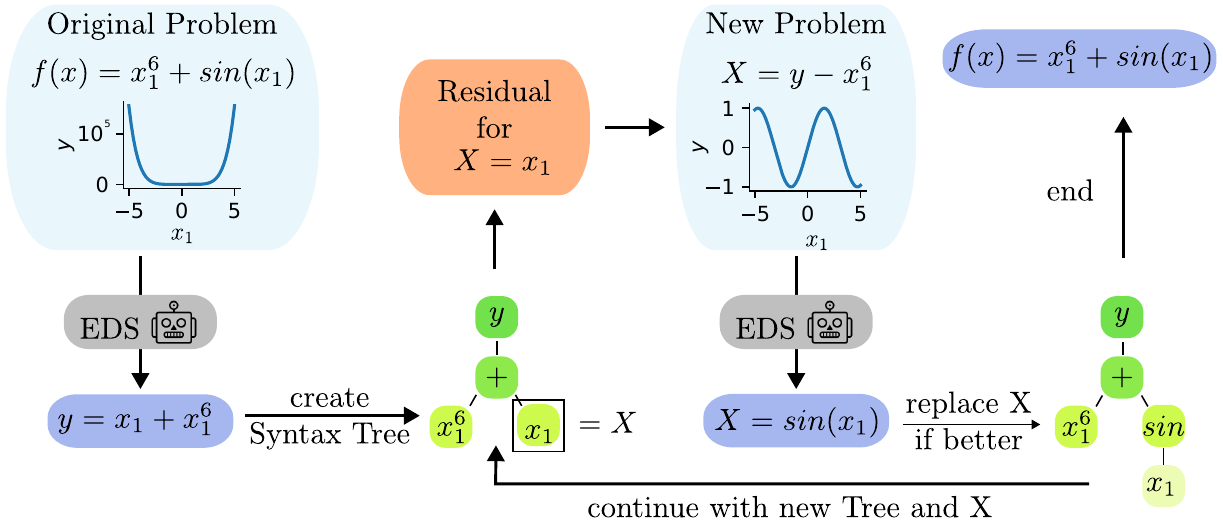}
\caption{
Overview of how RED helps to disentangle the initial problem $f(x) = x_1^6 + \sin(x_1) $. RED can be combined with every Equation Discovery System (EDS).
}
\label{fig:residual_scheme_intro}
\end{center}
\end{figure}
\section{Introduction}

Equation discovery is the task given a data set $D \in \mathbb{R}^{m \times n} $ to find for $m$ examples a free-form equation $f(\mathbf{x}, \mathbf{c}) = \mathbf{y}$ mapping the independent variables $x_i | 1 \leq i \leq n-1$ and the constants $\mathbf{c}$ to the dependent variable $y$. To find this equation, pretrained equation discovery systems (EDSs) have become popular in recent years  \cite{biggio_neural_2021,kamienny_end--end_2022,kamienny_deep_2023,shojaee_transformer-based_2023,udrescu_ai_2020,valipour_symbolicgpt_2021}. These systems use a neural architecture to embed the data set and are trained to predict the equation that generated the data set in a zero-shot way. 
The method proposed in this paper, 
\emph{Residuals for equation discovery (\ours{})}, calculates and optimizes the residuals of the data set for a subequation $X$ of the initial equation; that is, we compute what that subequation should have yielded for each data point such that the entire formula predicts the output correctly.
These residuals $\mathbf{y'}$  formulate a new problem $f'(\mathbf{x}, \mathbf{c}) = \mathbf{y}'$, and the equation discovery system can predict a solution. If the new solution's error is lower than that of the old solution, the new solution can replace $X$ in the original equation. 
In \cref{fig:residual_scheme_intro}, \ours{} is applied to the example equation $f(x) = x_1^6 + sin(x_1) $. 
\ours {} can help EDSs find a better solution after an initial prediction for two reasons: First, it can find a new representation of the original problem. Neural-guided EDSs are pretrained on many example equations and data sets. The chance of finding a representation similar to a previously seen problem increases by generating different representations of the original task with the residuals. Second, it can disentangle the original problem and divide it into simpler problems. The initial equation may already contain parts of the solution. 
By calculating the residual, the correct part of the initial equation is factored out of the original problem, leaving the unsolved subproblem for the EDS to address directly. Based on this, we address the following research questions in this work: How can the residuals be calculated without high  computational overhead (\textbf{R1})? Can \ours{} help equation discovery systems to find better equations (\textbf{R2})? How and when should \ours{} be applied to refine an equation (\textbf{R3})? What are the limitations of \ours{} (\textbf{R4})?

\section{Method}\label{sec:method}
To explain how \ours{} works, we first introduce the concept of syntax trees to represent an equation $f$. Subsequently, we use the syntax tree to formulate an equation system with only one operation per equation. %
Rearranging the equation system towards a specific node $X$ gives us the residual for the node.
Next, we answer \textbf{R1} and show how to calculate the residual given a syntax tree directly by defining the behavior of an operator node based on which adjacent node is calling the operator node. Finally, we introduce our method, \ours{}, which uses residuals to generate a new prompt for an EDS and recursively improve an initial equation.

\subsubsection{Syntax Tree as Equation System} 
\begin{figure}[t]
\begin{center}
\includegraphics[width=\linewidth]{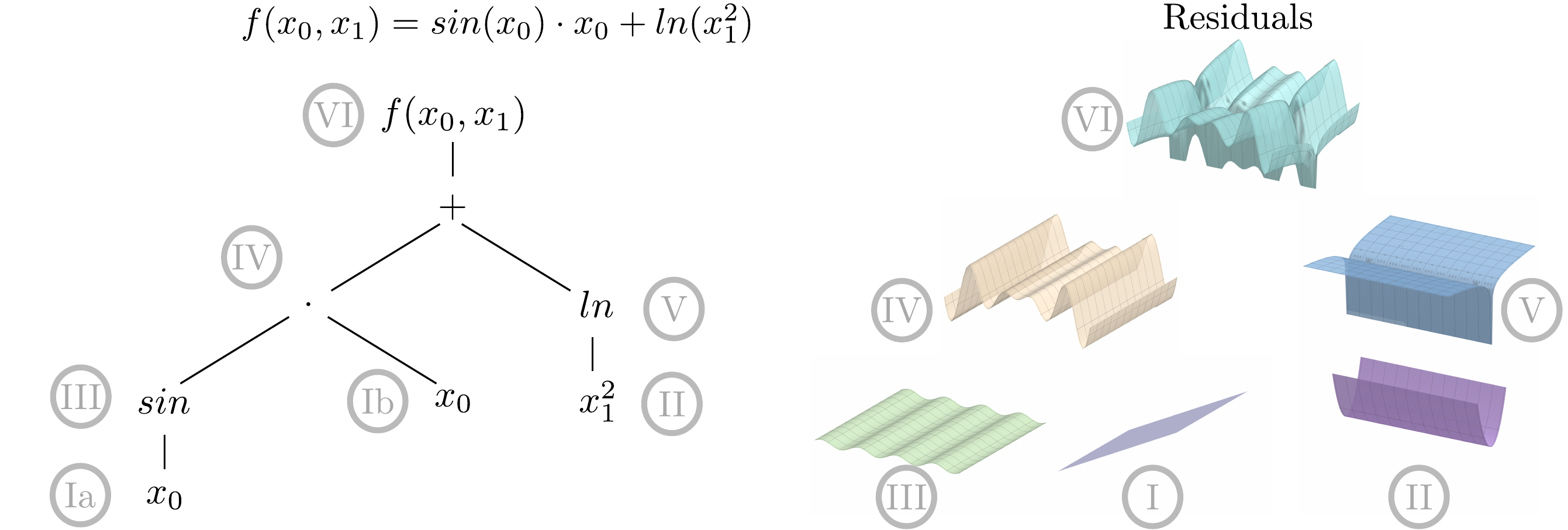}
\caption{
\textbf{left} Example for a syntax tree.  \textbf{right} Plots of the subequations composed to $f(x_0,x_1) = sin(x_0) \cdot x_0
 + ln(x_1^2)$
}
\label{fig:example_syntax_tree}
\end{center}
\end{figure}

To represent an equation, we use a syntax tree in which the inner nodes are operators and functions, and the leaf nodes are variables and constants. If a leaf node is called, it returns the corresponding value or the column from the data set.   Formally, a syntax tree \( \mathcal{S} \) consists of nodes \( \mathcal{N} \) and edges \( \mathcal{E} \). 
The edges connect two nodes $n_i$ and $n_j $ where $ n_i , n_j \in \mathcal{N} \wedge i \not = j$.
Unlike the usual definition of a syntax tree, we add an $Y$-node as a parent node to what would normally be the root node of the equation. The Y-node returns the y-column of the data set and is used to calculate the residuals.  
In a syntax tree, an equation is a composition of simple basic components as constants $\mathbf{c}$ and variables $\mathbf{x}$, which are combined into complex expressions by mathematical operations. In \cref{fig:example_syntax_tree},  one possible syntax tree for  $f(x_0,x_1) = sin(x_0) \cdot x_0 + ln(x_1^2)$ is given. For each node in the syntax tree, the equation equivalent to the corresponding subtree is plotted.
Using the example $f(x_0,x_1) = sin(x_0) \cdot x_0 + ln (x_1^2)$, we show how a syntax tree can be translated into an equation system in \cref{equ:equation_system}. 
We use the Roman numerals from \cref{fig:example_syntax_tree} to represent the result of a subtree.
\begin{align}
\begin{split}
VI = IV + V   ,\quad 
IV = III * Ib ,\quad 
Ib = x_0, \quad 
III = \sin(x_0) ,\quad 
V = \ln(x_1^2).
\end{split}
\label{equ:equation_system}
\end{align}

The above equation system can be rearranged to any node and defines what the subequation below a node must result in, so that the total equation is equal to y. The result of the calculation we name $y_{res}$. The only requirement is that all operators on the path between the node for which  $y_{res}$ is searched and the root node are invertible.

\subsubsection{Calculating the Residual based on a Syntax Tree}\label{sec:residuals_with_syntax_tree}
To answer \textbf{R1}, we can use the idea of an equation system to calculate the residuals for any node in the syntax tree. 
For an operator node, the mathematical operation it performs depends on which adjacent node is calling it. An overview of the most essential operator nodes is in \cref{tab:behaviour_of_node_typ}. As the calculation is defined on the node level, new mathematical operations can be added as new node types. \\
If we want to calculate the residuals for a node it calls its parent node. The parent node calls its other adjacent nodes, and they propagate the call recursively. When the results of the adjacent nodes are returned to the parent node,  the parent node performs its defined mathematical operation and returns the result to the node that called it in the first place. 
Operators that are not bijective (e.g., $sin$) cannot be inverted. Thus, for their child nodes, the residual cannot be computed.
\begin{table} [t!]
\caption{The mathematical behavior of a node type depends on the caller node. N/A indicates that the inverse is not applicable, while NI indicates that a node is not invertible.}
\label{tab:behaviour_of_node_typ}
\centering
\scriptsize
\begin{tabularx}{0.99\linewidth}{XYYY|XYY}
Binary Ops & \multicolumn{3}{c|}{Caller} & Other Nodes & \multicolumn{2}{c}{Caller} \\
 & \makecell[c]{Parent ($p$)} & \makecell[c]{Child 0} & \makecell[c]{Child 1} & & \makecell[c]{Parent} & \makecell[c]{Child 0}\\ 
 & \makecell[c]{($p$)} &  \makecell[c]{($c_0$)} & \makecell[c]{($c_1$)} & & \makecell[c]{($p$)} &  \makecell[c]{($c_0$)}\\ 
\midrule
Plus & $c_0 + c_1$ & $p - c_1$ & $p - c_0$ & Logarithm & $ln(c_0)$ & $e^p$ \\
Minus & $c_0 - c_1$ & $p + c_1$ & $c_0 - p$ & Sine & $sin(c_0)$ & NI \\
Product    & $c_0 * c_1$ & $p / c_1$ & $p / c_0$ & Constant & Value  & N/A \\
Division & $c_0 / c_1$ & $p * c_1$ & $c_0 / p$ & Variable & $x_i$ & N/A\\
Power & $c_1^{c_0}$ & $ln(p) / ln(c_1) $ & $p^{1/c_0}$ & Y & N/A & $y$  \\
\end{tabularx}
\end{table}

\subsubsection{\ours{}}\label{sec:method_RED}
\begin{figure}
\begin{center}
\includegraphics[width=0.98\linewidth, 
]{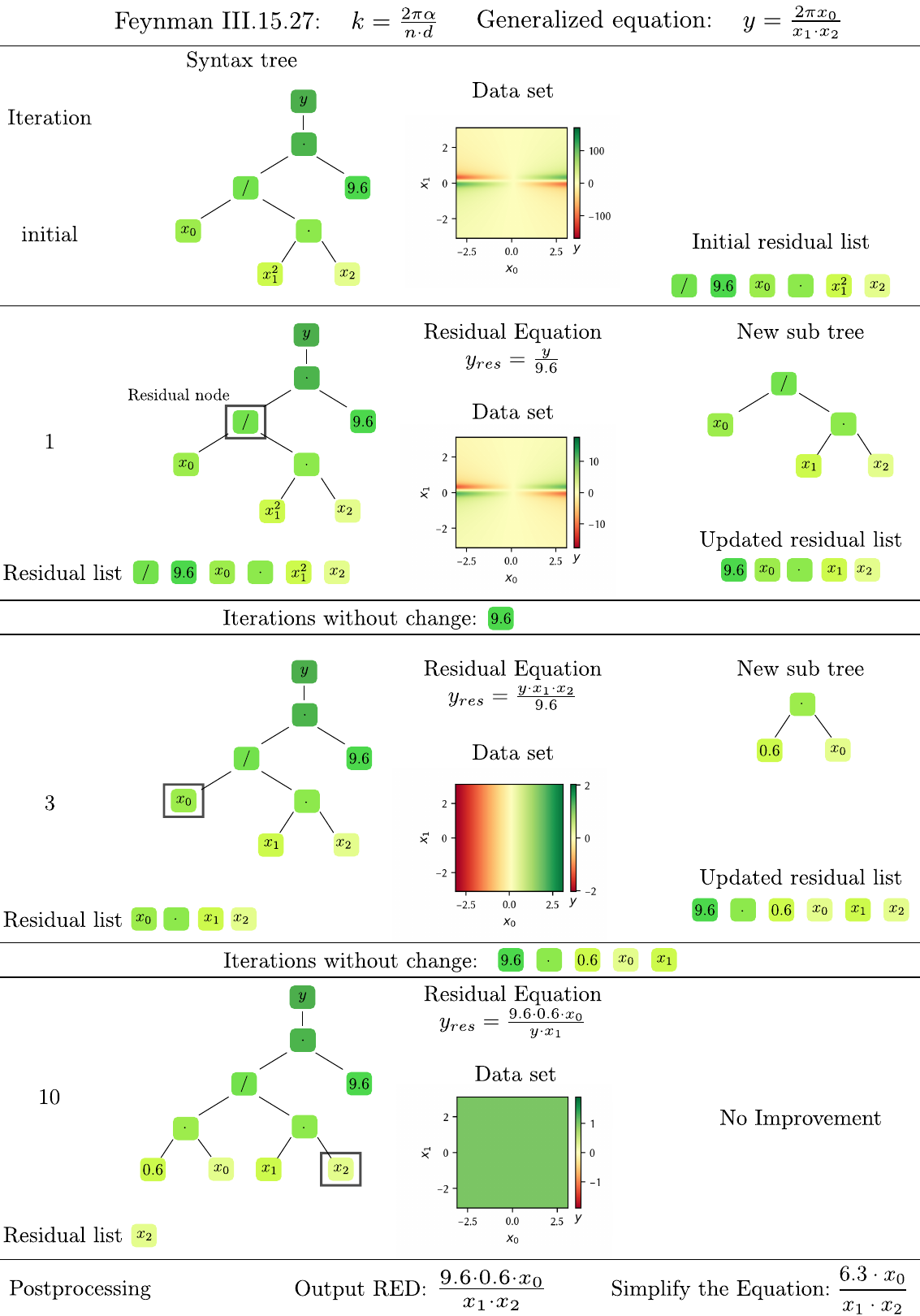}
\caption{
Example how \ours{} can help to find a better equation  using residuals in multiple iterations. For the visualization of the data set, $x_2$ is set to 1. 
}
\label{fig:residual_scheme}
\end{center}
\end{figure}

We can use the ability to calculate the residual for each node in a syntax tree to develop \ours{}. The pseudocode of \ours{} is given in \cref{alg:RED}. An example for applying \ours{} is shown in \cref{fig:residual_scheme}. 

\ours{} needs an initial equation $tree$, which we want to improve, and a current error $error^{val}$ as a threshold to check if the suggested new subequations improve the initial equation.
The idea for \ours{} is to calculate in an ordered manner for each node in the syntax tree whether the initial equation can be improved by using residuals. If a syntax tree is updated, \ours{} is restarted for the new tree until an iteration limit is reached or no further improvement is possible in the tree (line 3). 

The Residual List maintains the order of the nodes for which the residual is calculated. As in a breadth-first search, the ordering is from left to right within the nodes of the same depth and then goes to a level deeper. 
The child node of the Y-node is not added to the list, as using the residual for this node is identical to the original problem. 
The method $tree.\textsc{GetNextResidualNode}$ in line 5 returns the first node from the Residual List.
For the residual calculated in line 6, the EDS predicts a new subtree (line 7). This subtree replaces the current node in the original tree and creates a new tree, $tree_{res}$ (line 8). The error of $tree_{res}$ is calculated on the validation data set (line 9). 
If the error is smaller than the previous best error, we update the best error, and the tree to be optimized is set equal to $tree_{res}$ (line 11 - 12). Finally, $lastUpdatedNode$ in line 13 is set to the current $nodeId$.
 If $tree.GetNextResidualNode$ is called again in line 5, the Residual List for the new tree is created. Since we are traversing the tree from top to bottom, and only the tree below the $lastUpdatedNode$ has changed, we delete the $lastUpdatedNode$ and all its direct parent nodes from the Residual List. 
  For the removed nodes, the new subtree does not influence the calculation of their residual, and the problem is equivalent to a previously considered problem.

\begin{algorithm}[bt]
\caption{RED}\label{alg:search}
\begin{algorithmic}[1]
\Global Equation Discover Model $\mathcal{M}$, Train Data $\mathcal{D}_{train}$, Validation Data  $\mathcal{D}_{val}$ 
\Require Equation as Syntax Tree  $tree$, Residual Max Iteration $i_{max}$, Error Threshold $T$, Error of Current Equation $error^{val}$. 
\Function{\ours{}}{$tree$, $i_{max}$, T, $error^{val}$}
    \State $i, lastUpdatedNode \gets 0, 0$
    \While{$error^{val}$ > $T$ and $i$ < $i_{max}$ and \Call{len}{$tree.residualList$} > 0}
        \State $i \gets  i + 1$
        \State $nodeId \gets tree$\Call{.getNextResidualNode}{$lastUpdatedNode$}
        \State $res \gets tree$\Call{.getResidual}{$nodeId, \mathcal{D}_{train}$} \label{code:calculate_res}
        \State $subtree \gets \mathcal{M}$\Call{.fit}{$\mathcal{D}_{train}$, $res$} 
        \State $tree_{res} \gets$ \Call{updateSubTree}{$nodeId$, $tree$, $subtree$}
        \State $error^{val}_{res} \gets $\Call{testEquation}{$tree_{res}$, $\mathcal{D}_{val}$ }
        \If{$error^{val}_{res}$ < $error^{val}$}
        \State $error^{val}\gets error^{val}_{res}$
        \State $tree \gets tree_{res}$
        \State $lastUpdatedNode \gets nodeId$
        \EndIf
    \EndWhile
    \State \Return $tree$
\EndFunction
\end{algorithmic}
\label{alg:RED}
\end{algorithm}

\section{Experiments}\label{sec:experiments}
We compare\footnote{Code is available: \\ https://anonymous.4open.science/r/RED-Residuals-for-Equation-Discovery-02C3/} \ours{} for five EDSs with six other post-processing methods on the Feynman benchmark \cite{udrescu_ai_2020} as reported in SRBench \cite{Contemporary_Symbolic_Regression_Methods_and_their_Relative_Performance}. For \emph{NeSymReS}  \cite{biggio_neural_2021}, we use a pretrained model designed for a maximum of three independent variables. Therefore, we use data sets with three or fewer variables for all models. This results in 53 Feynman data sets. 
We first present the EDS models and post-processing methods in the following. Subsequently, we answer \textbf{R2} by comparing the performance of the post-processing methods. Finally, we examine \textbf{R3} and analyze how the number of iterations, noise, and data set size impact \ours{}.

\subsection{Models}
\emph{NeSymReS} \cite{biggio_neural_2021} uses  large-scale pretraining on generated data which are encoded with the Set Transformer architecture \cite{lee_set_2019}. A beam-search is used to sample candidates from the decoder. %
\emph{SymbolicGPT} \cite{valipour_symbolicgpt_2021} uses a transformer architecture too, but with a permutation-invariant data set encoder PointNet \cite{qi_pointnet_2016} which interprets the data sets as point clouds. 
\emph{E2E} \cite{kamienny_end--end_2022} is a transformer-based model directly predicting the full mathematical expression, constants included. In a subsequential step, the constants are refined by feeding them to an optimizer as an informed initialization. The data set is processed as sequential text. 
\emph{PySR} \cite{cranmer2023interpretable} is a genetic programming approach
tying the selection probability of the fittest individual (equation) to the increase in fit quality via simulated annealing. 
PySR outputs the discovered equations as a Pareto frontier, where for each complexity level  the equation with the least error is returned. In our experiment, we use the equation returned with the strategy ``best'', balancing equation complexity and error.
\emph{GPLearn} \cite{GP_learn} is an equation discovery library implementing genetic programming and is written in Python.

\subsection{Methods for post-processing Equations}\label{sec:Experiments:post processing method}

\begin{figure}[t]
\begin{center}
\includegraphics[width=\linewidth]{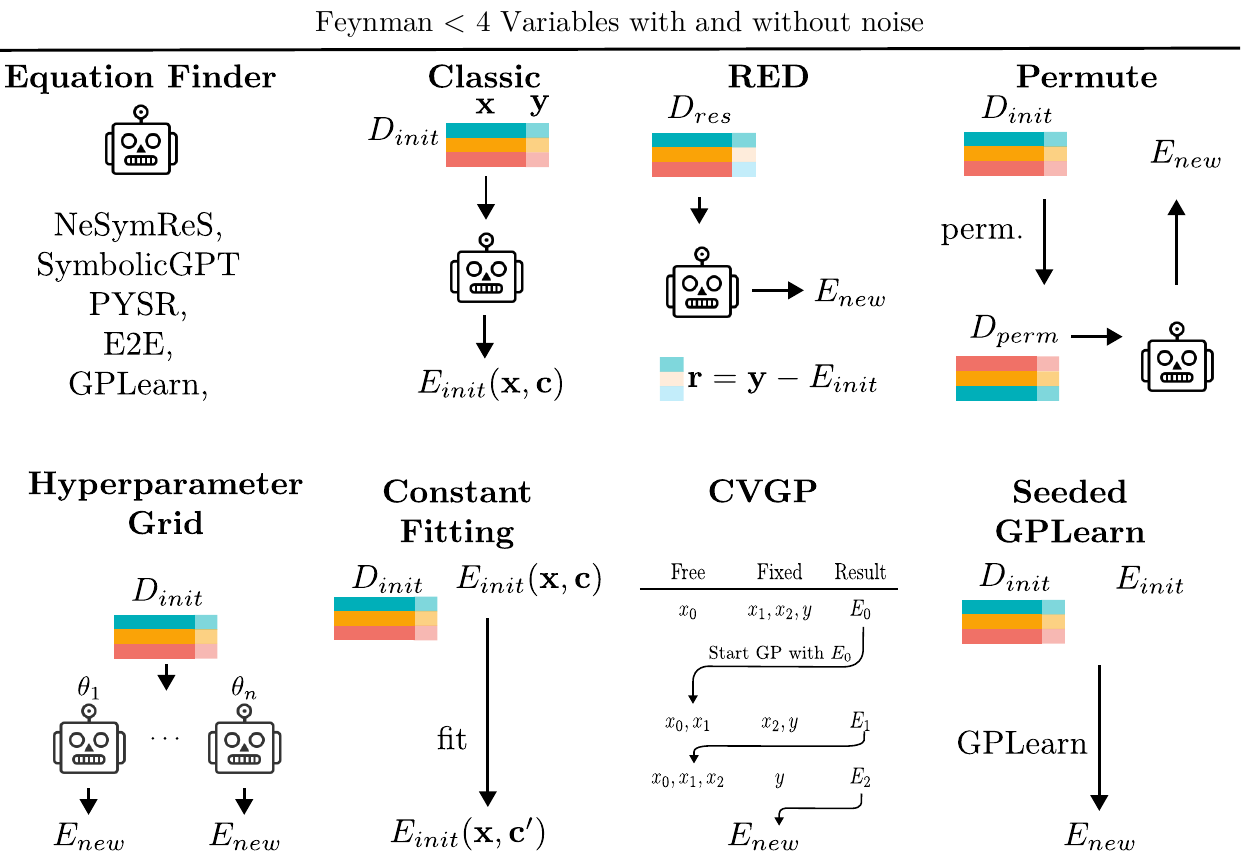}
\caption{
Overview of the tested post-processing methods.
}
\label{fig:overview_post_processing}

\end{center}
\end{figure}

In the following, we describe each post-processing method, and \cref{fig:overview_post_processing} gives a visual overview of the methods. 

\emph{\classic{}}: The EDS is applied once with the default parameter on the data set. The result is used as the initial equation for the other approaches.
\emph{\ours{}}: Described in detail in \cref{sec:method}. It uses residuals to refine an initial equation. Based on our experiments in \cref{sec:Experiment:How_often_residuals}, we select a maximum of 10  iterations to balance runtime, overfitting, and accuracy. 
\emph{\per{}}: The EDS is applied $n$-times on the permuted data set. Through the permutation, the EDS gets a new representation of the original data set with each permutation. $n$ is set equal to the performed iterations in \ours{}.  
\emph{\hyper{}}: In addition to the default parameter, two other sets of parameters are defined for each EDS. The sets differ in parameters like the probability of mutations in genetic approaches or the beam size in neural-guided EDSs.
\emph{\fitting{}}: All constants in the initial equation are refitted with the Levenberg-Marquardt algorithm from sklearn \cite{sklearn}. 
\emph{\cvgp{}}: Control Variable Genetic Programming %
\cite{cvgp} starts by fitting simple expressions involving a small set of independent variables with all other variables held constant. It then extends expressions learned in previous generations by adding new independent variables, using new control variable experiments in which these variables are allowed to vary. Applying control variable experiments can be interpreted as a post-processing method, which can be combined with every EDS as long the EDS can be seeded with an initial equation. For this reason, we group \cvgp{} as a post-processing method. To do this, we run \cvgp{} on all data sets of the benchmarks and then filter the results according to the data sets in which the current EDS has an error greater than 0.001. \cvgp{} needs the ground truth expression to perform the control variable experiments.
\emph{\seeded{}}: We replace 50\% of the 1000 candidates in the start population of GPLearn with a subtree of the initial equation. We start a run for each node in the syntax tree. Similar to \ours{}, the subtree is the complete tree without the subtree under the selected node.

We overall employ the following test protocol:
In the first phase, we run the \classic{} approach on each data set in the benchmark. If the Mean Squared Error (MSE) is greater than 0.001, we consider the data set as unsolved and test the post-processing methods. 
Unless otherwise specified, 300 rows are sampled from each data set. 
60\% of the data is used as the training set, 20\% as the validation set, and 20\% as the test set. If a post-processing method does not use a validation data set, the validation data set is added to the training data set.  Each equation proposed by a post-processing method is tested on the test data set. We use the MSE as the test metric. 

\subsection{Performance of \ours{}}

We apply each post-processing method as described in section \ref{sec:Experiments:post processing method}.
All post-processing methods, except Constant Fitting, Classic and CVGP, propose multiple equations. To evaluate \ours{} and answer \textbf{R2}, we run two types of experiments. First, we compare all equations the post-processing methods predict for one EDS. Next, we compare the best equations each method predicts for all five EDSs.  If a method can not return an equation, we use one found by \classic.

\subsubsection{Win-Loss Statistics}
\begin{figure}[t]
\begin{center}
\includegraphics[width=0.8\linewidth]{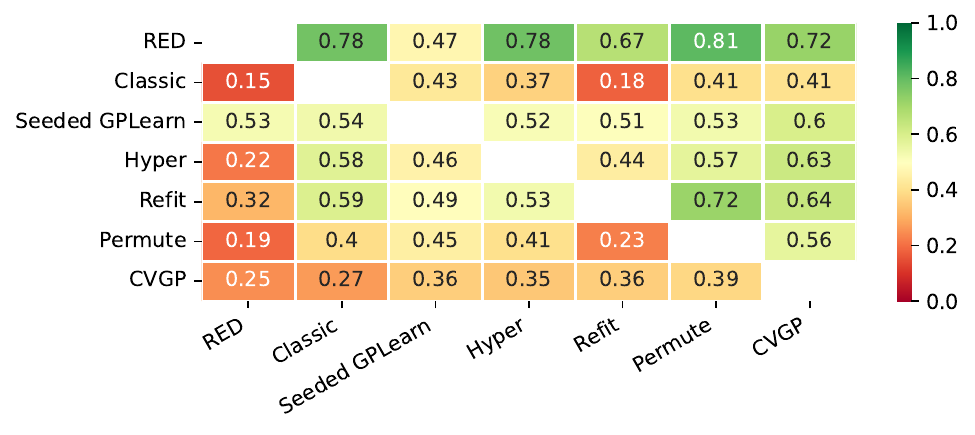}
\caption{
Win-Ratio for NeSymReS on the AI Feynman Benchmark. We compare two post-processing methods, one vs. one. For each data set we consider all equations both methods suggested and count how often one is better than the other. A ratio of 1 means that all equations of the method in the row are better than the equations of the method in the column  
}
\label{fig:win_loss}
\end{center}
\end{figure} 

 We compare for the EDS NeSymReS for each data set all equations a method predicted vs. the equations of another method. If the equation of the method in the row of \cref{fig:win_loss} has a lower MSE, we note a win, if equal a draw, and for a higher MSE a loss. In \cref{fig:win_loss} we present the   $\mathit{win}$-$\mathit{ratio}= \nicefrac{{wins}}{\mathit{wins}+\mathit{draws}+\mathit{loss}}$ for the AI Feynman benchmark. \Cref{fig:win_loss} demonstrates that \seeded{} wins more often than it loses against all methods. However, the difference compared to \ours{ } is only minimal. In contrast, the equations predicted by \ours{} are of higher quality than the other methods, resulting in a higher win ratio. 

\subsubsection{Accuracy for the best Run}
 In the following, we want to analyze the post-processing methods given the best equation they find for a data set. The results are given in  \cref{tab:Acc_sr_bench_results}.
 To reduce the proposed equations to one, we consider all proposed equations for a method and test them on the test data set. The method is then represented by the equation with the smallest error. 
First,  we summarize the results for each EDS (columns in the table) and afterward we review each post-processing method (rows).

For each EDS, we report the average \textbf{Running Time} to run all experiments over 3 seeds. This time is proportional to the time needed to query an EDS once and the length of the proposed equation. The length of the equation has an influence as the number of iterations of  \seeded{} and \ours{} depend on it. The metric \textbf{\# Completed} states for how many of the 53 equations the EDS returns an initial solution.

\begin{table}
\label{tab:Acc_sr_bench_results}
\caption{Results of the post-processing methods for multiple EDSs on the Feynman benchmark. We ran all experiments with 3 seeds, except E2E, which is only tested with one seed. For each EDS, the post-processing method with the lowest median is highlighted in \textbf{bold}. The methods whose performance is not significant ( $p \geq 0.01$) compared to \ours{} according to the Wilcoxon signed-rank test are marked with \underline{underline}.}
\scriptsize
\begin{tabularx}{\textwidth}{lRRRR}
\toprule
  &  NeSymReS  &  SymbolicGPT  &  PySR  &  GPLearn  \\ 
\midrule
Running Time [sec]  &  3735  &  4512  &  1823  &  12717  \\ 
Classic  &     &     &     &     \\ 
\qquad  \# Completed  &  52\phantom{.00}  &  33\phantom{.00}  &  53\phantom{.00}  &  46\phantom{.00}  \\ 
\qquad  \# MSE >0.001  &  22\phantom{.00}  &  32\phantom{.00}  &  22\phantom{.00}  &  31\phantom{.00}  \\ 
\qquad MSE Q2 ,  Q3  &  0.46  ,  10.24  &  \underline{1.88}  ,  28.58  &  0.05  ,  0.14  &  0.05  ,  0.99  \\ 
\qquad \# Operators Q2  &  4\phantom{.00}  &  6\phantom{.00}  &  2\phantom{.00}  &  3\phantom{.00}  \\ 
\qquad Runtime Q2 [sec]  &  3.27  &  1.67  &  1.93  &  17.55  \\ 
Seeded GPLearn  &     &     &     &     \\ 
\qquad  \# Completed  &  21\phantom{.00}  &  27\phantom{.00}  &  16\phantom{.00}  &  24\phantom{.00}  \\ 
\qquad MSE Q2 ,  Q3  &  \textbf{0.04}  ,  0.39  &  \textbf{0.02}  ,  0.10  &  \underline{0.02}  ,  0.25  &  0.05  ,  1.55  \\ 
\qquad \# Operators Q2  &  3\phantom{.00}  &  2\phantom{.00}  &  3\phantom{.00}  &  3\phantom{.00}  \\ 
\qquad Runtime Q2 [sec]  &  56.14  &  50.75  &  35.39  &  71.62  \\ 
Hyper  &     &     &     &     \\ 
\qquad  \# Completed  &  21\phantom{.00}  &  24\phantom{.00}  &  22\phantom{.00}  &  27\phantom{.00}  \\ 
\qquad MSE Q2 ,  Q3  &  \underline{0.21}  ,  1.88  &  \underline{2.48}  ,  20.53  &  \underline{0.02}  ,  0.15  &  \underline{\textbf{0.02}}  ,  0.26  \\ 
\qquad \# Operators Q2  &  5\phantom{.00}  &  10\phantom{.00}  &  2\phantom{.00}  &  3\phantom{.00}  \\ 
\qquad Runtime Q2 [sec]  &  7.86  &  4.17  &  3.72  &  34.82  \\ 
Refit  &     &     &     &     \\ 
\qquad  \# Completed  &  18\phantom{.00}  &  14\phantom{.00}  &  21\phantom{.00}  &  28\phantom{.00}  \\ 
\qquad MSE Q2 ,  Q3  &  0.29  ,  5.03  &  \underline{1.94}  ,  19.04  &  0.02  ,  0.11  &  \underline{0.03}  ,  0.36  \\ 
\qquad \# Operators Q2  &  4\phantom{.00}  &  4\phantom{.00}  &  2\phantom{.00}  &  2\phantom{.00}  \\ 
\qquad Runtime Q2 [sec]  &  0\phantom{.00}  &  0\phantom{.00}  &  0\phantom{.00}  &  0\phantom{.00}  \\ 
Permute  &     &     &     &     \\ 
\qquad  \# Completed  &  22\phantom{.00}  &  29\phantom{.00}  &  22\phantom{.00}  &  29\phantom{.00}  \\ 
\qquad MSE Q2 ,  Q3  &  0.46  ,  10.24  &  \underline{0.55}  ,  7.40  &  \underline{\textbf{0.01}}  ,  0.12  &  \underline{0.03}  ,  0.38  \\ 
\qquad \# Operators Q2  &  4\phantom{.00}  &  6\phantom{.00}  &  2\phantom{.00}  &  3\phantom{.00}  \\ 
\qquad Runtime Q2 [sec]  &  30.20  &  25.65  &  19.23  &  173.96  \\ 
CVGP  &     &     &     &     \\ 
\qquad  \# Completed  &  11\phantom{.00}  &  20\phantom{.00}  &  16\phantom{.00}  &  24\phantom{.00}  \\ 
\qquad MSE Q2 ,  Q3  &  \underline{5.94}  ,  22.84  &  \underline{0.15}  ,  9.03  &  5.61  ,  22.49  &  \underline{0.20}  ,  12.81  \\ 
\qquad \# Operators Q2  &  3\phantom{.00}  &  3\phantom{.00}  &  2\phantom{.00}  &  3\phantom{.00}  \\ 
\qquad Runtime Q2 [sec]  &  8.98  &  8.98  &  8.20  &  8.85  \\ 
\ours{}  &     &     &     &     \\ 
\qquad  \# Completed  &  14\phantom{.00}  &  19\phantom{.00}  &  15\phantom{.00}  &  18\phantom{.00}  \\ 
\qquad MSE Q2 ,  Q3  &  0.30  ,  5.46  &  1.85  ,  16.66  &  0.02  ,  0.12  &  \textbf{0.02}  ,  0.33  \\ 
\qquad \# Operators Q2  &  8\phantom{.00}  &  7\phantom{.00}  &  2\phantom{.00}  &  6\phantom{.00}  \\ 
\qquad Runtime Q2 [sec]  &  22.68  &  6.16  &  4.86  &  104.61  \\ 
\bottomrule
\end{tabularx}
\end{table}

Only PySR can provide a solution for all and NeSymReS for almost all of these equations. Both EDSs are also the fastest ones. The failures of the other EDSs can have several reasons: The neural-guided approaches are token-based. In the extreme case, these tokens are individual letters. Even if the approaches are trained to predict semantically sound formulas, this is not guaranteed. The EDSs sometimes predict expressions such as $sinI$. Another reason is that searching for a candidate equation results in over- or underflows, divisions by zero, undefined exponentiation in $\mathbb{R}$, or a negative logarithm. If the EDS or post-processing method does not handle the error, we log the error for the current process, and the program continues with the next method or data set. Finally, residuals cannot be applied if the EDS predicts an equation without an operator. The EDS E2E was tested with only one seed and then excluded from further experiments. Although this approach predicted equations with a small MSE,  the E2E has to be restarted multiple times, and has the longest suggested equations and running time. 

We now analyze the performance of the post-processing methods in detail. 
The metric \textbf{\# MSE$>$0.001} in \classic{} specifies the number of equations on which we test the post-processing methods. Each post-processing method has the metric \textbf{\# Completed}, which gives the number of data sets for which an equation is returned successfully. \textbf{MSE Q2} states the median of the mean squared error across all seeds and data sets. \textbf{MSE Q3} gives the third quartile correspondingly. 
To give an estimate of the complexity of the generated equations, \textbf{\# Operator Q2 } shows the median number of mathematical operators (e.g. $+$,$-$,$\cdot$, $/$, $\sin$, $\dots$) in the proposed equations. Finally, \textbf{Runtime Q2 [sec]} indicates the median duration each post-processing method ran per equation.

For \classic{}, the test error of the neural-guided EDSs are higher, and the equations found are more complex. The post-processing approach \seeded{} can close this gap and improve all examined EDSs. \ours{} also leads to an improvement, while for SymbolicGPT, it is not as drastic as for the other EDSs. One possible explanation is that \ours{} heavily depends on the initial equation, and the initial equations of  Symbolic GPT have the highest error. If this initial equation does not help to decompose the original problem, the residual problem can become even more challenging than the original problem. A drawback of \ours{} is that the average length of the equations increases significantly. We will analyze this issue in \cref{sec:limitations} in detail. \hyper{}, \fitting{}, and \per{} show promising results, especially with the genetic algorithms, but with NeSymReS, they only lead to medium improvements. \cvgp{} can only keep up with the other approaches regarding the median value for GP and is otherwise significantly worse. This is presumably because the control variable approach does not help much in simplifying the equation finding for the benchmark. The extension Racing-CVGP \cite{Jiang_Xue_2024} tries multiple permutations of the control variable order, which results in a significant improvement.

In conclusion, and with regard to \textbf{R3}, we observe that \ours{} helps improve the results of all the tested EDSs. The combination of neural-guided EDS with genetic EDS, as in \seeded, also shows promising results. However, it is less targeted, and requires much more computation time.

\subsection{Properties of \ours{}}
To answer \textbf{R3}, how and when \ours{} should be used, we first study the influence of iterations in RED and then consider the influence of noise and data set length on its performance.

\subsubsection{Effect of Maximum Number of Iteration on  \ours{} }\label{sec:Experiment:How_often_residuals}

\begin{figure}[t]
\centering 
\includegraphics[width=0.75\linewidth]{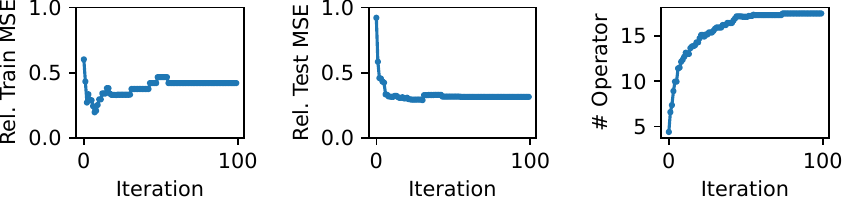}
\caption{
Influence of the number of iterations in \ours{} on the average relative MSE for train and test set and number of operators in the equations found with NeSymReS on the Feynman benchmark. 
}
\label{fig:average_error_iter_residuals}

\end{figure}

To analyze the effects of the number of iterations \ours{} has, we set the maximum number of iterations to 100. We store the train and test error and number of operations in the current equation per iteration for each data set. If we do not have the value for an iteration, we use the value from the last available iteration. 
We normalize the error for each data set so that the maximum error is 1. 
Subsequently, the mean value is calculated for all data sets per iteration and plotted in \cref{fig:average_error_iter_residuals}.
The figure shows that for NeSymReS (other EDSs behave similarly), the relative test error drops sharply within a few iterations and reaches a minimum. The average number of operators in the equation increases with the number of iterations.

\subsubsection{Effect of Noise on  \ours{} }
To quantify the effect of noisy data on \ours{}, a relative uniform noise ($U$) is added %
to each value in the data set $D$: $D_{noise} = D * (1 + U(- Rel. Noise, + Rel. Noise))$.

\begin{figure}[t]
\centering 
\includegraphics[width=0.8\linewidth]{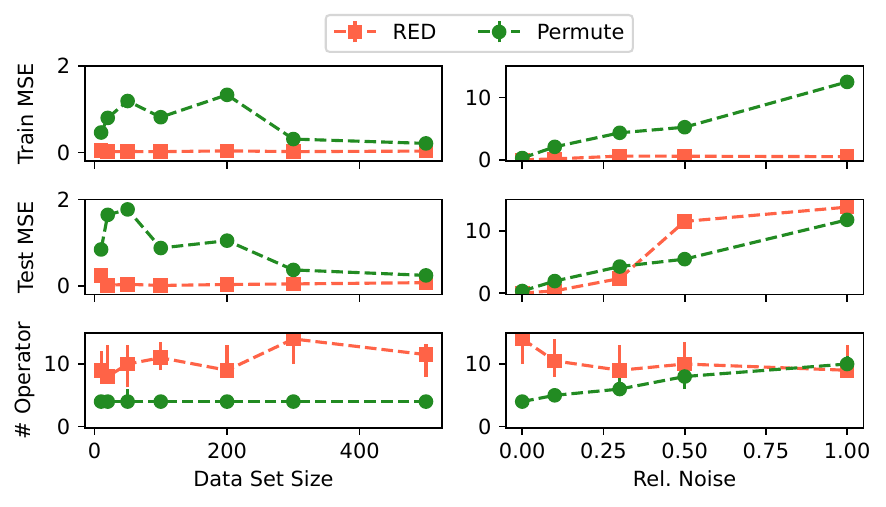}
\caption{
Influence of \textbf{data set size} (left) and \textbf{noise} (right) on the median number of operations, train, and test MSE for NeSymReS on the Feynman benchmark. 
}
\label{fig:rel_noise}
\end{figure}

For $D_{noise}$, the EDS NeSymReS is used to search for suitable equations with the approaches \per{} and \ours{}, and the best equation for each approach is logged for train and test error. 
For the Feynman benchmark, the median values for train and test MSE error, number of operations in the test equation, and number of successful fits are given as a function of the relative noise 0, 0.1, 0.3, 0.5, and 1 in \cref{fig:rel_noise}. 
The evaluation shows that the training error for the  \per{} approach increases linearly with noise. The training error for \ours{} remains almost constant despite the increase in noise.
This overfitting on the training data leads to similar values on the test data as with the \per{} approach. It is interesting to note that increasing noise leads to a decrease in the median number of operations for \ours{}, whereas an increase can be observed for \per{}.

\subsubsection{Effect of Data Set Size on \ours{}}

The number of data points available for the search can be another source of overfitting. We plot the median values for train, test error, and number of operations in \cref{fig:rel_noise} for the Feynman benchmark. We run the experiments for data sets of size 10, 20, 50, 100, 200, 300 and 500. 
For \per{}, we see the train and test error decrease with an increasing number of data points. The inability of NeSymRes to fit a few data points is surprising and is probably because the neural model was not trained for such small data sets. However, combined with \ours{}, the MSE remains at a constant low level regardless of the number of samples.

\subsubsection{Conclusion: Properties of \ours{}} To summarize our sensitivity analysis, we can state for \textbf{R3} that \ours{} can already improve the results with a small number of iterations. In particular, it can strongly improve the results in cases where the EDS has not seen a lot of training data. From a certain level of noise on, \ours{} tends to overfit. 

\section{Limitations}\label{sec:limitations}
The conducted experiments show that \ours{} is a promising method to improve the results of any EDS. Nevertheless, to answer \textbf{R4}, we want to summarize the challenges here and discuss possible solutions.

First, we address the challenges of  \ours{} with neural-guided EDSs and then with genetic EDSs. 
While \ours{} is independent of the functionality of the pretrained EDS, it depends on an initial solution, which has to enable the disentanglement. The initial solutions from SymbolicGPT have the highest median error of all EDSs, and accordingly, the results for \ours{} are less substantial than for NeSymReS. Further improvements for neural-guided EDSs can be expected, in which case \ours{} can be even more effective. 
The second challenge is that \ours{} leads to longer and thus more complex equations. 
We want to improve the search in future research to tackle this issue. Using not only the validation error but also the length of a new subequation as a criterion if the subequation is accepted will attenuate the tendency of \ours{} towards long equations. A parallel search combined with a dynamic ranking will achieve improvements within the same running time.

In the experiments with genetic EDSs, we show that \ours{} can achieve improvements there. Since genetic approaches perform a random search, the advantage of decomposing the original problem is not so evident here. One possible explanation for the good results of \ours{} with genetic EDSs, is that \ours{} effectively increases the number of search iterations for a data set, as a previously found equation can be further improved. However, an experiment in which we consider the median error of PySR depending on the number of iterations on the Feynman benchmark
shows no decrease in error with more iterations. 
Another explanation is that we restart the genetic algorithm multiple times within \ours{}; by resetting the hyperparameter, we find a solution outside the local maximum, which is not possible by simply extending the runtime of the genetic algorithm. The mutual influence of genetic EDSs and \ours{} and using our method as an operator in genetic algorithms is a promising approach for the future.

\section{Related Work}
  
Brute force approaches in equation discovery quickly reach their limits because the search space increases exponentially as the number of variables or operators increases. Since the beginning, equation discovery systems have been using metrics and methods to structure the search space. 

The most prominent method is to exploit the reward of a suggested equation as a metric to concentrate the search in promising areas. This procedure can be applied with genetic programming \cite{koza1994genetic}\cite{burlacu_operon_2020}, reinforcement learning \cite{petersen2021deep}, \cite{kamienny_deep_2023} or large language models \cite{shojaee2025llmsr}. 

An orthogonal approach tries to modify the data set that describes the problem and makes it easier to solve. \emph{BACON} \cite{langley_scientific_1987} adds new variables to the data set based on predefined rules to make the problem solvable by a simple combination of the new set of variables. 
\emph{CVGP} \cite{cvgp} starts with a few control variables and refines the initial solution by iteratively taking more variables into account. 
Similar to our work, because the target variable changes, is \emph{additive regression}. While \ours{} makes no assumption about the underlying structure, additive regression  \cite{breiman1985estimating} uses an additional superposition of continuous single-variable functions. The next term added to the existing one is fitted to minimize the error between the target variable and the equation so far found. 

\ours{} tries to decompose the original problem by considering subtrees of the initial solution. \emph{AI Feynman} \cite{udrescu_ai_2020} and  \emph{AI Feynman 2.0} \cite{udrescu_ai_2020-1} follow the same idea by training a neural network on the data set and then using the gradients of this network to decompose the problem into smaller sub-problems whose solutions are combined to a final equation. 

Generating different representations of a given problem to get better results from a pretrained neural system is already successfully applied in other domains. In the visual domain, Test-Time Augmentation (TTA) is used to create multiple versions of an image by, e.g., cropping and making decisions based on an ensemble \cite{kim2020learning}. In prompt engineering for LLMs \cite{Wang2023PlanandSolvePI,liu2023pre}, we iteratively refine the prompt for querying the LLM to produce multiple answers. 

\section{Conclusion}
In this research, we proposed \ours{}, a method using residuals to refine an initial equation. Given a syntax tree, we showed that we can calculate the residuals as fast as evaluating the tree using node-based calculation rules (\textbf{R1}). In experiments with two benchmarks, we demonstrated that \ours{} improves all equation discovery systems that we tested (\textbf{R2}). In a sensitivity analysis, we showed that 10 iterations of \ours{} already led to significantly improved results, and \ours{} helped in cases where the neural EDS has not seen a lot of training data (\textbf{R3}). 
Finally, we gave an outlook on how the increase in equation complexity when using \ours{} can be countered and how the approach can be further improved in the future by an optimized search strategy (\textbf{R4}).   

\begin{credits}
\subsubsection{\ackname} This research project was partly funded by the Hessian Ministry of Science and the Arts (HMWK) within the projects ``The Third Wave of Artificial Intelligence - 3AI'' and hessian.AI.

\subsubsection{\discintname}
\small{
The authors have no relevant financial or non-financial interests to disclose.
}

\subsubsection{Version of Record}
\small{
This preprint has  not undergone peer review or any post-submission improvements or 
corrections. The Version of Record of this contribution is published in  Lecture Notes in Computer Science, volume 16090, and is available online at:\\ \href{https:doi.org/10.1007/978-3-032-05461-6_7}{https:doi.org/10.1007/978-3-032-05461-6\_7}.
}
\end{credits}

\bibliographystyle{splncs04}
\bibliography{library4}

\clearpage

\end{document}